\pdfoutput=1

\documentclass[11pt]{article}

\usepackage{acl}

\usepackage{times}
\usepackage{latexsym}

\usepackage[T1]{fontenc}

\usepackage[utf8]{inputenc}

\usepackage{microtype}
\usepackage[nolist]{acronym}
\usepackage{multirow}
\usepackage[nolist]{acronym}
\usepackage{subcaption}
\usepackage{graphicx}
\usepackage[mathscr]{euscript}
\usepackage{amsmath}
\usepackage{amssymb}
\DeclareSymbolFont{rsfs}{U}{rsfs}{m}{n}
\DeclareSymbolFontAlphabet{\mathscrsfs}{rsfs}
\usepackage{algorithm}
\usepackage[noend]{algpseudocode}
\usepackage{adjustbox,booktabs}
\usepackage{hvfloat}

\definecolor{custom_orange}{RGB}{225, 165, 56}
\definecolor{custom_blue}{RGB}{0, 118, 186}
\definecolor{custom_green}{RGB}{87, 175, 28}

\title{Efficient Few-shot Learning for Multi-label Classification \\ of Scientific Documents with Many Classes}

\author{Tim Schopf, Alexander Blatzheim, Nektarios Machner, and Florian Matthes \\
         Technical University of Munich, Department of Computer Science, Germany\\ \texttt{\{tim.schopf,alexander.blatzheim,nektarios.machner,matthes\}}\\
         \texttt{@tum.de}}
         
\begin{document}

\begin{acronym}
\acro{nlp}[NLP]{Natural Language Processing}
\acro{icl}[ICL]{In-Context Learning}
\acro{pet}[PET]{Pattern-Exploiting Training}
\acro{peft}[PEFT]{Parameter Efficient Fine-Tuning}
\acro{plm}[PLM]{Pre-trained Language Model}
\acro{dare}[DARE]{Drop And Rescale}
\acro{llm}[LLM]{Large Language Model}
\acroplural{llm}[LLMs]{Large Language Models}
\acro{slerp}[SLERP]{Spherical Linear Interpolation}
\acro{le}[LE]{Label Embedding}
\end{acronym}

\maketitle
\begin{abstract}
Scientific document classification is a critical task and often involves many classes. However, collecting human-labeled data for many classes is expensive and usually leads to label-scarce scenarios. Moreover, recent work has shown that sentence embedding model fine-tuning for few-shot classification is efficient, robust, and effective. In this work, we propose FusionSent (\textbf{Fusion}-based \textbf{Sent}ence Embedding Fine-tuning), an efficient and prompt-free approach for few-shot classification of scientific documents with many classes. FusionSent uses available training examples and their respective label texts to contrastively fine-tune two different sentence embedding models. Afterward, the parameters of both fine-tuned models are fused to combine the complementary knowledge from the separate fine-tuning steps into a single model. Finally, the resulting sentence embedding model is frozen to embed the training instances, which are then used as input features to train a classification head. Our experiments show that FusionSent significantly outperforms strong baselines by an average of $6.0$ $F_{1}$ points across multiple scientific document classification datasets. In addition, we introduce a new dataset for multi-label classification of scientific documents, which contains 203,961 scientific articles and 130 classes from the arXiv category taxonomy. Code and data are available at \href{https://github.com/sebischair/FusionSent}{https://github.com/sebischair/FusionSent}.
\end{abstract}

\section{Introduction}

 \begin{figure*}[ht!]
    \centering
    \resizebox{1.0\linewidth}{!}{%
    \includegraphics{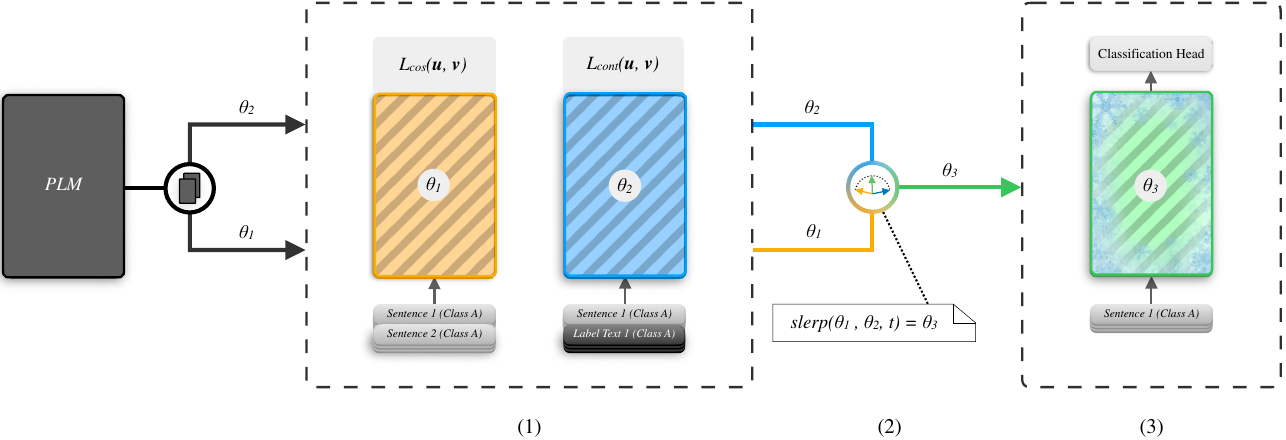}}
    \caption{The training process of FusionSent comprises three steps: (1) Fine-tune two different sentence embedding models from the same Pre-trained Language Model (PLM), with parameters \textcolor{custom_orange}{$\theta_{1}$}, \textcolor{custom_blue}{$\theta_{2}$} respectively. \textcolor{custom_orange}{$\theta_{1}$} is fine-tuned on pairs of training sentences using cosine similarity loss and \textcolor{custom_blue}{$\theta_{2}$} is fine-tuned on pairs of training sentences and their corresponding label texts, using contrastive loss. Label texts can consist of simple label/class names or of more extensive texts that semantically describe the meaning of a label/class. (2) Merge parameter sets \textcolor{custom_orange}{$\theta_{1}$}, \textcolor{custom_blue}{$\theta_{2}$} into \textcolor{custom_green}{$\theta_{3}$} using Spherical Linear Interpolation (SLERP). (3) Freeze \textcolor{custom_green}{$\theta_{3}$} to embed the training sentences, which are then used as input features to train a classification head.}
    \label{fig:fusionsent}
\end{figure*}

Scientific literature has grown exponentially over the last few decades, with countless new publications being added every year \cite{10.1145/3097983.3098016}. To be searchable and accessible to researchers, policymakers, and the public, scientific literature must be managed and categorized in digital libraries \cite{toney-dunham-2022-multi}. However, this poses a significant challenge due to the huge volume of documents and the variety of topics they cover \cite{sadat-caragea-2022-hierarchical}. In addition to the broad spectrum of possible topics, scientific documents often cannot be assigned to just one topic due to their interdisciplinary character. Consequently, automatically categorizing scientific documents must be approached as a multi-label classification problem over large label spaces. Previous works approach this task either in an unsupervised \cite{shen-etal-2018-web,10.1007/978-3-030-30760-8_26,mustafa_2021,toney-dunham-2022-multi,schopf-matthes-2024-nlp} or in a fully supervised \cite{10.1145/3487553.3524677,sadat-caragea-2022-hierarchical,e-mendoza-etal-2022-benchmark,schopf-etal-2023-exploring} manner. While supervised approaches offer high prediction quality, they require a large corpus of annotated data to perform. Often, however, a large corpus of annotated data is unavailable, e.g., when a new categorization scheme is being developed for an emerging scientific field. Unsupervised approaches provide a possible circumvention of this limitation but are accompanied at the expense of prediction quality. 

To improve classification performance in scenarios where labeled data is unavailable, domain experts may annotate a small part of the dataset. 
However, annotating many classes naturally leads to data scarcity, as collecting sufficient training data for all classes causes significantly higher costs \cite{xu-etal-2023-dense}. 
Therefore, to support the classification of scientific documents in such scenarios, we consider the multi-label classification of scientific documents as a few-shot task. Few-shot approaches are designed to train an effective model with a few labeled examples, reducing the cost of developing models for new domains and tasks \cite{huang-etal-2023-adasent}.

In recent work, SetFit \cite{tunstall2022efficient} demonstrated strong few-shot classification performance by contrastively fine-tuning \cite{Koch2015SiameseNN} sentence embedding models. Since this approach does not require prompts and is effective on relatively small models, it is much more efficient and consistent than common prompt-based methods such as \ac{icl} \cite{NEURIPS2020_1457c0d6} and \ac{pet} \cite{schick-schutze-2021-exploiting}, which involve careful prompt engineering and large-scale model sizes. 

In this paper, we propose \textit{FusionSent}, which builds on the idea of contrastive sentence embedding training for efficient few-shot classification. 
As illustrated in Figure \ref{fig:fusionsent}, FusionSent uses the few annotated examples, as well as label texts, to contrastively fine-tune two separate sentence embedding models from the same \ac{plm} checkpoint. One model is fine-tuned to maximize similarities between training examples sharing the same class, and the other model is fine-tuned to maximize similarities between training examples and their corresponding label texts. After fine-tuning, the weights of both models are merged to obtain the model body of FusionSent. For subsequent classifier training, the model body is frozen to embed the few training examples, which are then used as input features to train a simple logistic regression head. This approach works effectively with relatively small model sizes, requires no prompts, and merging fine-tuned sentence embedding models incurs no additional inference or memory costs \cite{pmlr-v162-wortsman22a}. 
Our experiments show that FusionSent consistently outperforms various baselines on different datasets for multi-label classification of scientific documents with many classes. Furthermore, we show that FusionSent can improve few-shot performance in multi-class settings of different domains with a small number of classes. 

In addition to FusionSent, we introduce a new dataset for multi-label classification of scientific documents. The dataset consists of 130 classes and 203,961 scientific articles that have been manually categorized by their authors into one or more topics from the arXiv category taxonomy\footnote{\href{https://arxiv.org/category_taxonomy}{https://arxiv.org/category\_taxonomy}}.

\section{Related Work}
\subsection{Classification of Scientific Documents}
Unsupervised approaches typically use embeddings of topics as well as scientific documents and perform classification based on their similarities \cite{shen-etal-2018-web,10.1007/978-3-030-30760-8_26,mustafa_2021,toney-dunham-2022-multi}. More recently, classifying scientific documents has been regarded as a fully supervised task. SciNoBo \cite{10.1145/3487553.3524677} uses the structural properties of publications and their citations and references organized in a multilayer graph network for predicting topics of scientific publications. HR-SciBERT \cite{sadat-caragea-2022-hierarchical} uses a multi-task learning approach for topic classification with keyword labeling as an auxiliary task. Finally, \citet{e-mendoza-etal-2022-benchmark} use ensemble models to classify scientific documents into multiple research themes.

\subsection{Few-shot Classification}
Prominent techniques for few-shot classification involve \ac{icl}, utilizing task-specific prompts with a few labeled examples \cite{NEURIPS2020_1457c0d6}. However, while avoiding gradient updates, \ac{icl} necessitates large model sizes for good performance, resulting in computationally expensive inference. Conversely, prompt-based fine-tuning proves to be effective with smaller models \cite{schick-schutze-2021-exploiting,tam-etal-2021-improving,gao-etal-2021-making}. Additionally, \ac{peft} can further reduce training costs by fine-tuning a considerably smaller module within a frozen \ac{plm} \cite{pmlr-v97-houlsby19a,li-liang-2021-prefix,hu2022lora,karimi-mahabadi-etal-2022-prompt,he2022towards,NEURIPS2022_0cde695b,aly-etal-2023-automated}. In contrast to these methods, fine-tuning few-shot classification models via contrastive sentence embedding training provides two primary advantages: (1) it requires significantly smaller model sizes, and (2) eliminates the necessity for prompts or instructions \cite{tunstall2022efficient,huang-etal-2023-adasent,bates-gurevych-2024-like}, which can cause significant performance variance and require careful design \cite{NEURIPS2021_5c049256}.


\subsection{Model Fusion}

Model fusion, which involves the integration of capabilities from different models, can be mainly divided into two categories. Firstly, ensemble approaches combine the output of multiple models to enhance the overall prediction performance \cite{LITTLESTONE1994212,https://doi.org/10.1002/widm.1249}. Outputs are typically combined by weight averaging \cite{LITTLESTONE1994212} or majority voting \cite{6033566}.
These ensemble approaches can improve the prediction performance of large-scale language models \cite{jiang-etal-2023-llm}. Secondly, weight merging approaches enable model fusion at the parameter level. \citet{pmlr-v162-wortsman22a} show that weight averaging of multiple models fine-tuned with different hyperparameters improves prediction accuracy and robustness. Task vectors derived from model weights can be modified and combined together through arithmetic operations to steer the behavior of a resulting model \cite{ilharco2023editing}. This approach can be enhanced by trimming task vectors and resolving sign conflicts before merging them \cite{yadav2023tiesmerging}. In addition, \ac{dare} can be used as a general preprocessing technique for existing model merging methods to merge multiple task-specific fine-tuned models into a single model with diverse abilities \cite{yu2023language}.

\subsection{Datasets for Topic Classification of Scientific Documents}
Various datasets for multi-label topic classification of scientific documents have been introduced. The Cora dataset \cite{10.1023/A:1009953814988} contains about 50,000 computer science research papers categorized into 79 topics. Several datasets have been released based on the ACM Computing Classification System\footnote{\href{https://dl.acm.org/ccs}{https://dl.acm.org/ccs}} \cite{santos2009multi,sadat-caragea-2022-hierarchical}. \citet{schopf-etal-2023-exploring} introduce a dataset of 179,349 scientific papers categorized into 82 different NLP-related topics. \citet{yang-etal-2018-sgm} create a dataset of 55,840 arXiv\footnote{\href{https://arxiv.org}{https://arxiv.org}} papers, in which each paper is assigned to several classes covering 54 different topics. However, this dataset is not publicly available. 

\section{Background}

\paragraph{Sentence Embedding Model Fine-tuning for Few-shot Classification} 
\citet{tunstall2022efficient} show that sentence embedding models can be used in a two-step training process for efficient few-shot classification. In the first step, a sentence embedding model is fine-tuned in a contrastive manner by sampling positive and negative sentence pairs from few-shot labeled examples. 
In the second step, the fine-tuned sentence embedding model is frozen to encode all available few-shot examples. The resulting embeddings are then used as input features to train a simple logistic regression classifier \cite{abd18569-0124-3b1e-9ca9-67a6fd857a26} as the model head. 

\paragraph{Label Texts for Document Classification in Label-scarce Scenarios} 
\citet{xu-etal-2023-dense} show that mapping representation spaces of training instances to their respective label descriptions in embedding space can be effective in label-scarce classification scenarios. They reformulate classification with many classes as a dense retrieval task and train a dual encoder that learns to maximize the similarity between embeddings of the training instances and their respective label descriptions. 
During inference, they use the top-$k$ retrieved labels of each instance for classification. Similarly, WanDeR \cite{10.1145/3539618.3592085} and FastFit \cite{yehudai2024llms} use label names and dense retrieval for multi-class classification. However, dense retrieval approaches are challenging to apply in multi-label classification scenarios since the number of classes per instance can vary significantly. 

\section{Method} \label{sec:fusionsent_method}

As illustrated in Figure \ref{fig:fusionsent}, our few-shot classification method consists of separate training parts for the model body and the model head. We fine-tune a sentence embedding model as the model body, while the model head consists of a simple logistic regression classifier trained on the data encoded by the model body.

\subsection{Model Body} \label{sec:model_body}

Given a base \ac{plm}, we fine-tune the model body of FusionSent in three steps: (1) use SetFit's contrastive learning approach to construct positive and negative training pairs from the few training examples to fine-tune a sentence embedding model from the base \ac{plm} (2) construct positive and negative training pairs from the few training examples and their corresponding label texts to fine-tune a different sentence embedding model using the same base \ac{plm}, and (3) merge both fine-tuned sentence embedding models to obtain the model body of FusionSent. 

In the first step, we fine-tune a sentence embedding model from the base \ac{plm} using contrastive learning and the few training examples. Specifically, from the few training examples, instances of the same class are selected as positive pairs, which are assigned a score of 1, and instances from different classes are selected as negative pairs, which are assigned a score of 0. These training pairs are then used to fine-tune a sentence embedding model with the Cosine Similarity Loss: 
\begin{equation}
    L_{cos}=\|y-cos\_sim(u,v)\|_{2},
\end{equation}

where $u,v \in \mathbb{R}^D$ are the $D$-dimensional sentence embeddings of two sentences respectively, and $y \in \{0,1\}$ is the pair label.

In the second step, we use a different contrastive training approach to fine-tune a separate sentence embedding model from the same base \ac{plm}, using the few training instances and their corresponding label texts. Specifically, positive pairs consist of training instances and the label texts of the class assigned to them. Negative pairs consist of training instances and label texts from different classes. Label texts can consist of simple label/class names, which are usually available in datasets, or of more extensive texts that semantically describe the meaning of a label/class. We assign the positive pairs a score of 1 and the negative pairs a score of 0 to fine-tune a sentence embedding model with the Contrastive Loss \cite{1640964}: \begin{multline}
L_{cont}=\frac{1}{2}\bigl[y\cdot cos\_dist(u,v)^{2}+\\(1-y)\cdot max\{0,m-cos\_dist(u,v)\}^{2}\bigr],
\end{multline}

where $u,v \in \mathbb{R}^D$ are the $D$-dimensional sentence embeddings of two sentences respectively, $m=0.5$ is a margin, and $y \in \{0,1\}$ is the pair label.

To obtain the contrastive training pairs for steps one and two, we use an \textit{oversampling} strategy. In this approach, an equal number of positive and negative training pairs are sampled, with the minority pair type (positive) being oversampled to align with the majority pair type (negative).

In the third step, the parameters of the fine-tuned sentence embedding models obtained in steps one and two are merged using \ac{slerp} \cite{10.1145/325334.325242}. Specifically, let $\theta_{1}$ be the parameters obtained from the first fine-tuning step and $\theta_{2}$ the parameters obtained from the second fine-tuning step, we merge parameters with \ac{slerp}:
\begin{equation}
    slerp(\theta_{1},\theta_{2};t)=\frac{\sin (1-t) \Omega}{\sin \Omega} \theta_{1} +
\frac{\sin t \Omega}{\sin \Omega} \theta_{2},
\end{equation}

where $\theta_{1}\cdot\theta_{2} = \cos\Omega$ and $t=0.5$ is an interpolation factor. Finally, the new parameters $\theta_{3}$ obtained from \ac{slerp} merging are inserted into a sentence embedding model derived from the same architecture as the base \ac{plm}, resulting in the FusionSent model body.

\subsection{Model Head \& Inference}

In the second part of FusionSent training, we first use the frozen model body to embed all available training instances. Then, we train a logistic regression model using the embedded training instances as input features. During inference, the model body embeds the inputs to provide features for the logistic regression head that subsequently classifies the unseen instances.

\section{Experiments}
\subsection{Data}

We construct a dataset of scientific documents derived from arXiv metadata \cite{clement2019usearxivdataset}. The arXiv metadata provides information about more than 2 million scholarly articles published in arXiv from various scientific fields. We use this metadata to create a dataset of 203,961 titles and abstracts categorized into 130 different classes. To this end, we first perform a stratified downsampling of the metadata to only 10\% of all articles while retaining the original class distribution. Afterward, articles assigned to categories occurring less than 100 times in the downsampled dataset are removed. To obtain the final dataset, we then perform a stratified train/validation/test split of the processed dataset in an 80:10:10 ratio. The number of examples in each set are shown in Table \ref{tab:arxiv_data_summary}.

\begin{table}[h!]
    \centering
    \renewcommand{\arraystretch}{0.8} %
    \begin{tabular}{lc}
    \toprule
    \small{\textbf{Dataset split}} & \small{\textbf{Size}} \\
    \hline
    \small{Train}  &  \small{163,168} \\
    \small{Validation} & \small{20,396} \\
    \small{Test} & \small{20,397} \\
    \bottomrule
    \end{tabular}
    \caption{Overview of the arXiv dataset.}
    \label{tab:arxiv_data_summary}
\end{table}

\begin{figure}[h!]
    \centering
    \resizebox{1.0\columnwidth}{!}{%
    \includegraphics{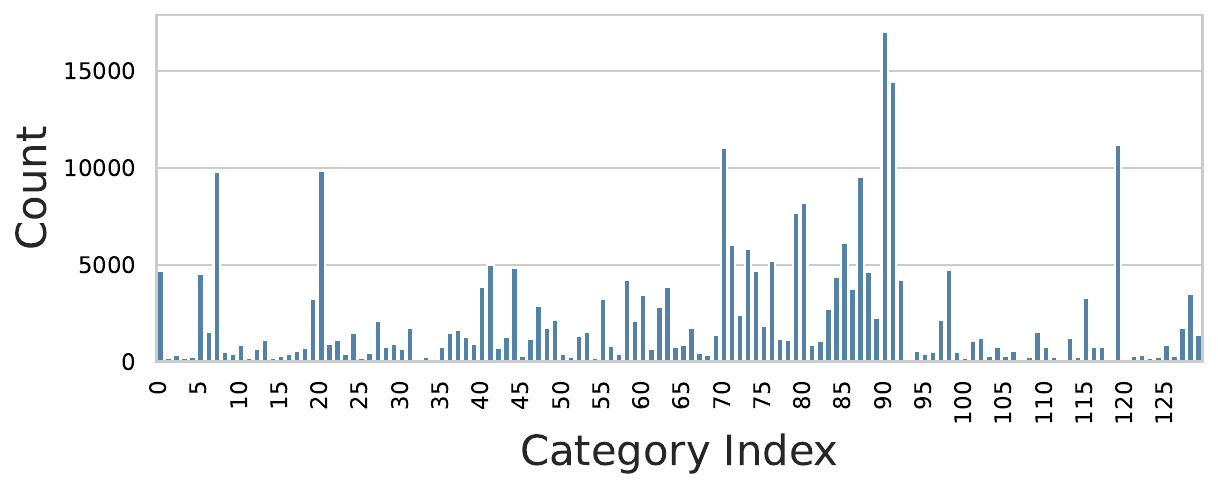}}
    \caption{Number of papers in each category of the arXiv dataset.}
    \label{fig:arxiv_data_distribution}
\end{figure}

Each article in the resulting arXiv dataset is categorized into one or more distinct categories. Figure \ref{fig:arxiv_data_distribution} shows the distribution of papers across the 130 categories of the dataset.

In addition, we use the SciHTC dataset \cite{sadat-caragea-2022-hierarchical}, which contains computer science papers categorized into one or more classes of the ACM Computing Classification System. We remove classes with less than 100 examples, resulting in 46,372 training samples and 5,838 test samples categorized into 62 different classes.

As a third dataset for scientific document classification, we use the NLP taxonomy dataset \cite{schopf-etal-2023-exploring}, which contains papers from the ACL Anthology\footnote{\href{https://aclanthology.org}{https://aclanthology.org}}, the arXiv cs.CL category, and Scopus\footnote{\href{https://www.scopus.com}{https://www.scopus.com}}, categorized into one or more \ac{nlp}-related classes. We perform a stratified 90:10 split between training and test examples, resulting in 161,414 training and 17,935 test instances categorized into 82 different classes.

\subsection{Models}

We experiment with two baselines and four different few-shot learning approaches for multi-label classification of scientific documents. 

\paragraph{FineTune} The first baseline consists of a standard encoder-only transformer that is fine-tuned for text classification.

\paragraph{SetFit} The second baseline consists of the SetFit approach without any changes to the architecture or the training procedure.

\paragraph{Label Embedding (LE)} As an initial few-shot learning approach, we experiment with only training one sentence embedding model that uses few-shot examples and their corresponding label texts in a contrastive learning approach. This approach consists of training a model body, as described in step 2 in Section \ref{sec:model_body}, and training a logistic regression head on top of it.

\paragraph{SetFit$\rightarrow$LE} We also experiment with combining contrastive learning approaches to directly fine-tune a single sentence embedding model rather than separate models that are merged later. For this purpose, we perform training steps one and two as described in Section \ref{sec:model_body} sequentially on the same sentence embedding model. For classification, we then train a logistic regression head.

\paragraph{LE$\rightarrow$SetFit} This approach also only trains a single sentence embedding model. However, we first perform training step two followed by step one as described in Section \ref{sec:model_body} on the same model. We then train a logistic regression model head for the classification. 

\paragraph{FusionSent} Finally, we experiment with the FusionSent approach as described in Section \ref{sec:fusionsent_method}.

\begin{table*}[ht!]
    \centering
    \resizebox{1.8\columnwidth}{!}{%
    \renewcommand{\arraystretch}{0.7} %
   
    \begin{adjustbox}{center}

    \begin{tabular}{l|ccc|ccc|ccc|c}
    \toprule
    {\textbf{Dataset $\rightarrow$}} & \multicolumn{3}{c}{\textit{arXiv}} & \multicolumn{3}{c}{\textit{SciHTC}} & \multicolumn{3}{c}{\textit{NLP Taxonomy}} & \multicolumn{1}{c}{\textit{Average}} \\

    \cmidrule(lr){2-4}
    \cmidrule(lr){5-7}
    \cmidrule(lr){8-10}
    \cmidrule(lr){11-11}
    
    {\textbf{Method $\downarrow$}} &  $\textbf{F}_{1}$ &  \textbf{P} &      \textbf{R} &    $\textbf{F}_{1}$  &   \textbf{P} &      \textbf{R} &    $\textbf{F}_{1}$  &     \textbf{P} &      \textbf{R} &    $\textbf{F}_{1}$ \\
    
    \midrule
    
    \multicolumn{11}{c}{$|N|=2^{\ast}$} \\
    {FineTune} & 
    - & - & - &  - & - & - & $13.1_{3.2}$ & $78.3_{5.5}$ & $7.2_{2.0}$ & -  \\
    {SetFit} & 
    $37.5_{1.6}$ & $45.5_{1.7}$ & $32.0_{2.4}$ & $31.4_{0.8}$ & $39.9_{0.1}$ & $26.0_{1.2}$ & $58.7_{2.0}$ & $55.7_{4.2}$ & $62.3_{0.9}$ & $42.5_{1.5}$ \\
    {Label Embedding} & 
    $43.2_{0.1}$ & $45.3_{1.9}$ & $40.4_{0.5}$ & $34.1_{4.3}$ & $46.1_{6.3}$ & $27.1_{3.3}$ & $65.8_{2.2}$ & $66.0_{5.5}$ & $66.0_{1.6}$ & $47.7_{2.2}$ \\
    {SetFit$\rightarrow$LE} & 
    $41.5_{0.9}$ & $45.3_{1.5}$ & $38.3_{1.7}$ & $32.3_{1.9}$ & $41.6_{3.2}$ & $26.3_{1.4}$ & $64.6_{1.8}$ & $62.7_{4.4}$ & $66.8_{2.1}$ & $46.1_{1.5}$ \\
    {LE$\rightarrow$SetFit} & 
    $39.2_{0.5}$ & $41.5_{1.1}$ & $37.1_{0.5}$ & $34.0_{1.6}$ & $38.9_{1.6}$ & $30.2_{2.0}$ & $61.3_{1.6}$ & $55.0_{3.4}$ & $69.3_{1.5}$ & $44.8_{1.2}$ \\
    {FusionSent} & 
    $\mathbf{44.4_{0.4}}$ & $50.5_{1.2}$ & $39.6_{0.1}$ & $\mathbf{36.7_{1.9}}$ & $48.0_{4.2}$ & $29.7_{0.9}$ & $\mathbf{66.2_{2.1}}$ & $67.4_{4.6}$ & $65.2_{0.8}$ & $\mathbf{49.1_{1.5}}$ \\
    \hline
    \multicolumn{11}{c}{$|N|=4^{\ast}$} \\
    {FineTune} & 
    - & - & - & $10.3_{1.6}$ & $59.3_{3.6}$ & $5.6_{0.9}$ & $43.5_{1.8}$ & $85.2_{2.1}$ & $29.2_{1.9}$ & - \\
    {SetFit} & 
    $45.6_{1.0}$ & $46.4_{1.6}$ & $45.1_{0.9}$ & $35.0_{0.3}$ & $42.6_{1.1}$ & $29.7_{0.3}$ & $63.6_{1.5}$ & $60.7_{2.3}$ & $66.9_{1.8}$ & $48.1_{0.9}$ \\
    {Label Embedding} & 
    $47.0_{1.1}$ & $47.7_{0.6}$ & $46.3_{2.0}$ & $32.7_{1.1}$ & $41.9_{2.3}$ & $26.9_{0.6}$ & $71.9_{0.6}$ & $70.8_{0.5}$ & $73.0_{1.2}$ & $50.5_{0.9}$ \\
    {SetFit$\rightarrow$LE} & 
    $46.0_{0.3}$ & $47.4_{0.4}$ & $44.7_{0.7}$ & $30.7_{2.4}$ & $36.9_{4.4}$ & $29.5_{2.1}$ & $70.0_{0.6}$ & $68.1_{0.8}$ & $72.1_{0.4}$ & $48.9_{1.7}$ \\
    {LE$\rightarrow$SetFit} & 
    $45.8_{1.0}$ & $45.0_{1.4}$ & $46.7_{0.9}$ & $32.0_{2.9}$ & $40.6_{2.3}$ & $27.8_{1.9}$ & $66.1_{0.7}$ & $62.1_{0.8}$ & $70.7_{0.9}$ & $48.0_{1.5}$ \\
    {FusionSent} & 
    $\mathbf{48.3_{1.1}}$ & $51.0_{1.0}$ & $46.0_{1.5}$ & $\mathbf{38.5_{2.3}}$ & $45.4_{1.0}$ & $33.5_{2.9}$ & $\mathbf{72.6_{0.5}}$ & $72.1_{0.3}$ & $73.2_{0.7}$ & $\mathbf{53.1_{1.3}}$ \\
    \hline
    \multicolumn{11}{c}{$|N|=8^{\ast}$} \\
    {FineTune} & 
    $18.7_{2.8}$ & $72.5_{2.4}$ & $10.8_{1.9}$ & $26.6_{3.9}$ & $55.0_{3.7}$ & $17.5_{3.1}$ & $67.1_{1.1}$ & $88.0_{1.6}$ & $54.2_{1.1}$ & $37.5_{2.6}$ \\
    {SetFit} & 
    $46.0_{0.6}$ & $44.1_{0.5}$ & $48.1_{1.1}$ & $37.5_{3.7}$ & $46.2_{3.2}$ & $31.5_{3.8}$ & $66.2_{0.2}$ & $66.4_{0.4}$ & $66.0_{0.4}$ & $49.9_{1.5}$ \\
    {Label Embedding} & 
    $47.8_{1.4}$ & $46.3_{1.5}$ & $49.5_{1.4}$ & $36.4_{3.2}$ & $41.3_{2.5}$ & $32.5_{3.5}$ & $77.2_{0.5}$ & $74.0_{0.7}$ & $80.8_{0.7}$ & $53.8_{1.7}$ \\
    {SetFit$\rightarrow$LE} & 
    $45.3_{1.2}$ & $45.5_{1.0}$ & $45.1_{1.4}$ & $30.5_{1.6}$ & $31.4_{2.0}$ & $29.7_{1.9}$ & $72.7_{0.4}$ & $70.5_{0.5}$ & $75.1_{0.4}$ & $49.5_{1.1}$ \\
    {LE$\rightarrow$SetFit} & 
    $44.1_{1.0}$ & $42.1_{1.8}$ & $46.3_{0.5}$ & $32.9_{3.0}$ & $41.8_{4.1}$ & $27.1_{2.3}$ & $66.7_{0.4}$ & $65.8_{1.3}$ & $67.6_{1.1}$ & $47.9_{1.5}$ \\
    {FusionSent} & 
    $\mathbf{49.0_{1.3}}$ & $49.0_{1.8}$ & $49.0_{0.9}$ & $\mathbf{41.2_{4.6}}$ & $43.4_{5.4}$ & $39.2_{3.9} $ & $\mathbf{78.3_{0.3}}$ & $76.0_{0.2}$ & $80.7_{0.5}$ & $\mathbf{56.2_{2.1}}$ \\
    \hline
    \hline
    \multicolumn{11}{c}{$|N|=Full^{\ast\ast}$} \\
    {FineTune} & 
    71.6 & 78.2 & 66.1 & 57.9 & 73.5 & 47.8 & 95.9 & 96.2 & 95.7 & 75.1 \\
    \bottomrule
    \end{tabular}
\end{adjustbox}%
}
\caption{FusionSent performance scores and standard deviations for few-shot classification of scientific documents compared to different approaches across three test datasets and four training set sizes $|N|$. Micro $F_{1}$,  Precision (P), and Recall (R) scores are reported. $^{\ast}$Number of training samples per class. $^{\ast\ast}$Entire available training data used. In some cases, insufficient training examples were provided for the FineTune model to learn, resulting in no class predictions during testing.} 
\label{tab:main-evaluation}
\end{table*}

\subsection{Experimental Setup} \label{sec:main-experimental-setup}
Systematically evaluating few-shot performance is challenging due to the potential instability arising from fine-tuning on small datasets \cite{zhang2021revisiting}. In our multi-label scientific document classification experiments, we use three random training splits for each dataset and sample size to mitigate this issue. For each method, we report the average measure and the standard deviation across these splits. We use SciNCL \cite{ostendorff-etal-2022-neighborhood} as the base \ac{plm} for each model. While we train the FineTune model for 50 epochs in the few-shot setting, we use the same approach to train a model on the full training datasets for 3 epochs. In both cases, we use a batch size of 12. 
For sentence embedding model training according to step 1 in Section \ref{sec:model_body}, we use a batch size of 4 for all models, and for training according to step 2 in Section \ref{sec:model_body}, we use a batch size of 1. Both steps are trained for 1 epoch for all models. In addition, each model is trained with a learning rate of $2e^{-5}$.

For the arXiv dataset, we use the publicly available category descriptions as label texts. For the SciHTC and NLP taxonomy datasets, we generate short descriptive texts from the provided label names with GPT-4 \cite{openai2023gpt4} and use them as label texts. Table \ref{tab:label_descriptions} shows examples of the used label names and label descriptions.

\section{Results}

Table \ref{tab:main-evaluation} shows a comparison between FusionSent and the other few-shot approaches for $|N|\in\{2,4,8\}$ labeled training samples per class. We observe that FusionSent consistently outperforms $F_{1}$ scores of all approaches investigated for each dataset and training set size. Further, FusionSent significantly outperforms SetFit across all training set sizes by an average of 6.0 $F_{1}$ points. While the other approaches using label texts for sentence embedding training on a single model can perform better than SetFit, they fall short of the FusionSent approach. The \ac{le} approach shows consistent improvements over SetFit on average, while the SetFit$\rightarrow$LE and LE$\rightarrow$SetFit approaches only outperform SetFit in a few cases.

The results demonstrate that using label texts for sentence embedding training can help to separate instances of different classes in the embedding space, providing a crucial property for the classification head to perform well. However, combining the contrastive sentence embedding training approaches of SetFit and \ac{le} in a single model does not significantly increase performance. Using a two-step contrastive training approach does not enable a single sentence embedding model to effectively encode information from both training steps, as it may suffer from forgetting previously acquired knowledge \cite{biesialska-etal-2020-continual}. Conversely, the FusionSent results indicate that this limitation can be circumvented by training separate sentence embedding models with different contrastive learning approaches and subsequently merging their parameters. This approach ensures that the individually trained models encode different information, whereas merging allows their respective knowledge to complement each other, resulting in improved model performance.

\section{Experiments with Few Classes} \label{sec:few-class-experiments}

To determine the generalizability of FusionSent to few-shot settings with a low number of classes and different domains, we perform experiments on the SST-5 \cite{socher-etal-2013-recursive}, CR \cite{10.1145/1014052.1014073}, Emotion \cite{saravia-etal-2018-carer}, AGNews \cite{NIPS2015_250cf8b5}, and EnronSpam \cite{metsis2006spam} datasets as processed for few-shot classification by \citet{tunstall2022efficient}. These datasets each comprise two to six classes and cover the fields of sentiment classification, opinion detection from customer reviews, emotion detection from \textit{Twitter} tweets, news article classification, and e-mail spam detection. For all datasets, we generate short descriptive texts from the provided label names with GPT-4 and use them as label texts. Table \ref{tab:label_descriptions} shows examples of the used label names and label descriptions. We experiment with all ten available randomized training splits for each dataset and sample size using $|N|=8$ and $|N|=64$ few-shot examples. We report the average accuracy and standard deviation across the training splits for each method.

We use the paraphrase-mpnet-base-v2\footnote{\href{https://huggingface.co/sentence-transformers/paraphrase-mpnet-base-v2}{https://huggingface.co/sentence-transformers/paraphrase-mpnet-base-v2}} model \cite{reimers-gurevych-2019-sentence} as base \ac{plm} for SetFit and FusionSent. Additionally, we use RoBERTa$_{LARGE}$ \cite{liu2019roberta} for the FineTune baseline. The other training parameters remain the same as in Section \ref{sec:main-experimental-setup}.

\begin{table}[t!]
    \centering
    \resizebox{1.0\columnwidth}{!}{%
    \renewcommand{\arraystretch}{1.0} %
    \begin{tabular}{l ccccccc}
    \toprule
    \textbf{Dataset $\rightarrow$} & \textbf{SST-5} & \textbf{AGNews} & \textbf{Emotion} & \textbf{EnronSpam} & \textbf{CR} & \multirow{2}{*}{\textbf{Average}}\\
    \cmidrule(lr){2-4}
    \cmidrule(lr){5-6}
    \textbf{Method $\downarrow$} & \multicolumn{3}{c}{Multi-class Classification} & \multicolumn{2}{c}{Binary Classification} & \\
    \midrule
    \multicolumn{7}{c}{$|N|=8^{\ast}$} \\
    FineTune$^{\dagger}$ & $33.5_{2.1}$ & $81.7_{3.8}$ & $28.7_{6.8}$ & $85.0_{6.0}$ & $58.8_{6.3}$ & $57.5_{5.0}$ \\
    SetFit & $41.7_{2.0}$ & $82.6_{3.6}$ & $49.5_{3.8}$ & $91.0_{3.2}$ & $89.6_{1.2}$ & $70.9_{2.8}$ \\
    FusionSent & $\mathbf{43.0_{3.2}}$ & $\mathbf{84.4_{2.2}}$ & $\mathbf{57.1_{2.5}}$ & $\mathbf{91.4_{3.9}}$ & $\mathbf{89.8_{1.0}}$ & $\mathbf{73.1_{2.6}}$ \\
    \hline
    \multicolumn{7}{c}{$|N|=64^{\ast}$} \\
    FineTune$^{\dagger}$ & $45.9_{6.9}$ & $88.4_{0.9}$ & $65.0_{17.2}$ & $95.9_{0.8}$ & $88.9_{1.9}$ & $76.8_{5.5}$ \\
    SetFit & $48.1_{4.4}$ & $87.7_{0.8}$ & $78.5_{2.0}$ & $96.1_{0.5}$ & $90.6_{0.7}$ & $80.2_{1.7}$ \\
    FusionSent & $\mathbf{50.0_{2.8}}$ & $\mathbf{88.3_{0.8}}$ & $\mathbf{78.7_{1.6}}$ & $\mathbf{96.5_{0.6}}$ & $\mathbf{90.7_{0.6}}$ & $\mathbf{80.8_{1.3}}$ \\
    \hline
    \hline
    \multicolumn{7}{c}{$|N|=Full^{\ast}$} \\
    FineTune$^{\dagger}$ & $59.8$ & $93.8$ & $92.6$ & $99.0$ & $92.4$ & $87.5$ \\
    \bottomrule
    \end{tabular}}
    \caption{FusionSent accuracy scores and standard deviations for few-shot classification with few classes compared to the baselines across five test datasets for three training set sizes. $^{\ast}$Number of training samples per class. $^{\ast\ast}$Entire available training data used. $^\dagger$Results from \citet{tunstall2022efficient}. }
    \label{tab:setfit_benchmark}
\end{table}

\paragraph{Results} Table \ref{tab:setfit_benchmark} shows the results of SetFit and FusionSent training on the binary and multi-class datasets. On average, FusionSent outperforms SetFit for $|N|=8$ by an average of 2.2 accuracy points. However, as the number of training samples increases to $|N|=64$, the gap decreases to 0.6 accuracy points. In addition, the improvements are more substantial for multi-class classification, whereas they are only minimal in the binary case. 

For binary classification, instances of different classes must be pushed apart from each other as far as possible to allow the logistic regression classifier to find a good decision boundary. For this relatively simple problem, the results indicate that SetFit can already effectively separate instances of different classes in embedding space. For multiple classes, however, it is more difficult to find positions in the embedding space that separate instances of different classes from each other. In these cases in particular, FusionSent shows its strengths and helps the classification head to find good decision boundaries between classes. 


\section{Robustness Against Label Text Variations}

To evaluate the robustness of FusionSent against different label text variations, we conduct experiments on the previous datasets using simple label names instead of extensive label descriptions. In these experiments, we simply use the label names as provided by the respective datasets and compare the classification results with those obtained by using detailed label descriptions. 
Table \ref{tab:label_descriptions} shows examples of the used label names and label descriptions. We use $|N|=8$ few-shot examples and report the average $F_{1}$ performance over the respective training splits. Furthermore, we use the same training parameters as in Section \ref{sec:main-experimental-setup} and Section \ref{sec:few-class-experiments}.

\begin{figure}[ht!]
    \centering
    \resizebox{1.0\columnwidth}{!}{%
    \includegraphics{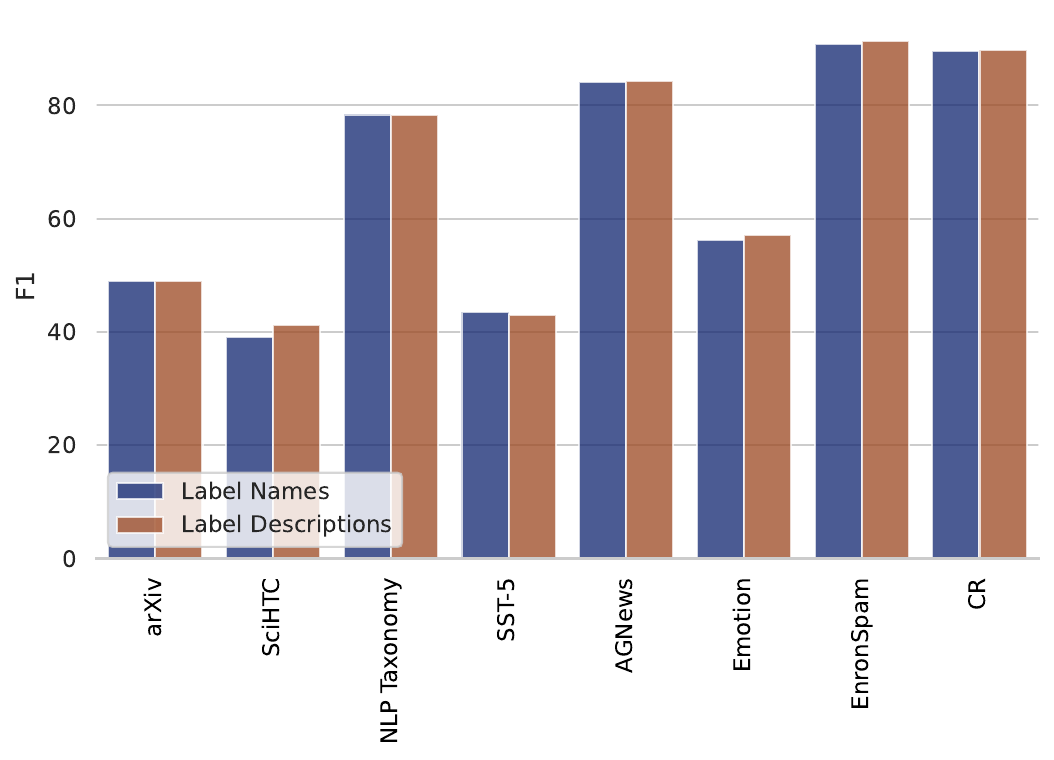}}
    \caption{FusionSent micro $F_{1}$ scores for few-shot classification on 8 different datasets using either extensive label descriptions or simple label names. We report the average score over the random training splits of each dataset using $|N|=8$ training examples per class.}
    \label{fig:label_names}
\end{figure}

\paragraph{Results} Figure \ref{fig:label_names} shows the performance of FusionSent using simple label names, as provided by the respective datasets, compared to using extensive textual label descriptions generated by GPT-4. We obtain similar performances across the different label text variants with a mean performance difference of $0.48$ and a standard deviation of $0.39$ $F_{1}$ points. Furthermore, there is no clear pattern as to whether the use of extensive label descriptions or simple label names leads to significantly improved performance. In comparison, performance variations of 10 accuracy points and more when using different prompts on the same model are characteristic for prompt-based few-shot classification approaches \cite{NEURIPS2021_5c049256}. Therefore, we conclude that FusionSent is relatively robust to label text variations and the use of simple label names is already sufficient to achieve good classification performance.

\section{Computational Costs}

To compare the relative computational costs of FusionSent and SetFit, we follow the approach of \citet{NEURIPS2022_0cde695b} and use FLOPs-per-token estimates, which can be obtained from \citet{kaplan2020scaling}. Specifically, encoder-only models with $N$ parameters have approximately $2N$ FLOPs-per-token for inference and $6N$ FLOPs-per-token for training. The resulting cost for inference and training is then given by:\begin{align}
   & C_{inf}=2N\cdot\ell_{seq},\\ 
   & C_{train}=6N\cdot\ell_{seq}\cdot\ n_{steps}\cdot n_{batch},
\end{align}

where $\ell_{seq}$ is the input sequence length, $n_{steps}$ is the number of training steps, and $n_{batch}$ is the batch size. Since we are training two model bodies for FusionSent, we calculate the training costs for each model body separately and then add them up. For inference, we can use the formula as provided, since we only use one model body.

We estimate the costs using the scientific document classification datasets from Table \ref{tab:main-evaluation} and SciNCL as base \ac{plm} with $N=110M$ parameters. Based on the median number of tokens per instance in all datasets, we use $\ell_{seq}=194$ to estimate the costs for training approaches that do not use label texts. Since we perform inference on these instances, we also use this value to estimate the inference cost for all approaches. Taking into account the shorter label texts, we use $\ell_{seq}=130$ to estimate the costs for training approaches that utilize label texts. Additionally, we use fixed values of $n_{steps}=1,000$, and $n_{batch}=8$ for all training estimates.
\begin{table}[h!]
    \centering
    \resizebox{1.0\columnwidth}{!}{%
    \renewcommand{\arraystretch}{1.0} %
    \begin{tabular}{lccc}
    \toprule
    \textbf{Method} & \textbf{Inf. FLOPs} & \textbf{Train FLOPs} & \textbf{Avg.} $\mathbf{F_{1}}$\\
    \hline
    SetFit & $4.3e10$ & $1.0e15$ & $49.9_{1.5}$ \\
    FusionSent & $4.3e10$ & $1.7e15$ & $56.2_{2.1}$ \\
    \bottomrule
    \end{tabular}}
    \caption{Computational costs and average micro $F_{1}$ scores of FusionSent and SetFit using $|N| = 8$ training samples on the scientific document classification datasets listed in Table \ref{tab:main-evaluation}.}
    \label{tab:computational_costs}
\end{table}

As shown in Table \ref{tab:computational_costs}, the increase in $F_{1}$ performance is accompanied by increased training costs. This is the result of training two sentence embedding models instead of one. However, by merging the models, the inference efficiency remains the same as when using the base \ac{plm}.  Although FusionSent incurs higher training costs, it can significantly improve prediction performance while maintaining SetFit's inference efficiency. 

\section{Conclusion}

We introduce FusionSent, a new approach for efficient and prompt-free few-shot classification of scientific documents. FusionSent uses label texts and contrastive learning to improve classification performances over several other few-shot approaches. We show that FusionSent is particularly effective in scenarios with many classes while being computationally efficient during inference. Additionally, FusionSent is robust against label text variations. Finally, we introduce a new arXiv dataset for multi-label classification of scientific documents.





\bibliography{anthology,custom}
\bibliographystyle{acl_natbib}

\clearpage
\appendix
\twocolumn[\section*{Appendix}
\label{sec:appendix}

\vskip\intextsep
\noindent\begin{minipage}{\textwidth}
    \centering
    \resizebox{1.0\linewidth}{!}{%
    \begin{tabular}{l|p{7cm}|p{10cm}}
    \toprule
    \textbf{Dataset} & \textbf{Label Names} & \textbf{Label Descriptions}\\
    \hline
     arXiv & General Relativity and Quantum Cosmology & General relativity and quantum cosmology focuses on gravitational physics, including experiments and observations related to the detection and interpretation of gravitational waves, experimental tests of gravitational theories, computational general relativity, relativistic astrophysics, solutions to Einstein's equations and their properties, alternative theories of gravity, classical and quantum cosmology, and quantum gravity. \\
      & ... & ... \\
    \hline
     SciHTC & Information retrieval & The "Information Retrieval" class within the 2012 ACM Computing Classification System encompasses the study and design of systems for indexing, searching, and retrieving information from large datasets. It includes the development of algorithms and techniques for processing and querying textual and multimedia data, as well as evaluating the effectiveness of retrieval systems. Key topics within this class involve search engine architectures, query representation, relevance feedback, and information extraction. The field also addresses challenges such as handling unstructured data, understanding user context, and ensuring privacy and security in the retrieval process. \\
     & ... & ...  \\
    \hline
     NLP Taxonomy & Named Entity Recognition & Named Entity Recognition is the identification and classification of entities (e.g., names of people, organizations) in text. \\
     & ... & ... \\
    \hline
    SST-5 & very positive & 'very positive' is used for data samples that express strong or intense positive sentiments, enthusiasm, or approval. \\
     & ... & ... \\
    \hline
    AGNews & Sports & 'Sports' represents data samples related to sports news, events, scores, and athlete performances. \\
     & ... & ... \\
    \hline
    Emotion & sadness & 'sadness' is characterized by feelings of hopelessness, disappointment, melancholy, and vulnerability, often accompanied by a sense of isolation or being overwhelmed. \\
     & ... & ... \\
    \hline
    EnronSpam & spam & 'spam' is an unsolicited and often irrelevant or inappropriate message sent over the internet, typically to a large number of users, for the purpose of advertising, phishing, spreading malware, or other malicious activities. \\
     & ... & ... \\
    \hline
    CR & negative & 'negative' corresponds to criticisms, complaints, or expressions of dissatisfaction with products or services. \\
     & ... & ... \\
    \bottomrule
    \end{tabular}}
    \captionof{table}{Examples of label names and extensive label descriptions for different datasets.}
    \label{tab:label_descriptions}
\end{minipage}\par\vskip\intextsep]

\end{document}